\documentclass[sigconf]{acmart}
\usepackage{enumitem}

\setcopyright{acmlicensed}
\copyrightyear{2018}
\acmYear{2018}
\acmDOI{XXXXXXX.XXXXXXX}
\acmConference[Conference acronym 'XX]{Make sure to enter the correct
  conference title from your rights confirmation email}{June 03--05,
  2018}{Woodstock, NY}
\acmISBN{978-1-4503-XXXX-X/2018/06}

\begin{document}
\title{Topic Modeling and Link-Prediction for \\Material Property Discovery}

\author{Ryan C. Barron}
\email{barron@lanl.gov}
\orcid{0009-0005-5045-9527}
\affiliation{%
    \institution{Theoretical Division, \\ Los Alamos National Laboratory}
  \city{Los Alamos}
  \state{New Mexico}
  \country{USA}
}
\affiliation{%
  \institution{CSEE, University of Maryland, Baltimore County }
  \city{Baltimore}
  \state{Maryland}
  \country{USA}
}
\author{Maksim E. Eren}
\affiliation{%
    \institution{Information Systems and Modeling, Los Alamos National Laboratory}
  \city{Los Alamos}
  \state{New Mexico}
  \country{USA}
}
 
\author{Valentin Stanev}
\affiliation{%
  \institution{Department of Material Science \& Engineering, University of Maryland }
  \city{ College Park}
  \state{Maryland}
  \country{USA}
}
\author{Cynthia Matuszek}
\affiliation{%
  \institution{CSEE, University of Maryland, Baltimore County }
  \city{Baltimore}
  \state{Maryland}
  \country{USA}
}
\author{Boian S. Alexandrov}
\affiliation{%
    \institution{Theoretical Division, \\ Los Alamos National Laboratory}
  \city{Los Alamos}
  \state{New Mexico}
  \country{USA}
}
\renewcommand{\shortauthors}{Barron et al.}
\begin{abstract}
Link prediction is a key network analysis technique that infers missing or future relations between nodes in a graph, based on observed patterns of connectivity. Scientific literature networks and knowledge graphs are typically large, sparse, and noisy, and often contain missing links, potential but unobserved connections, between concepts, entities, or methods. Here, we present an AI-driven hierarchical link prediction framework that integrates matrix factorization and human-in-the-loop visualization to infer hidden associations and steer discovery in complex material domains. Our method combines Hierarchical Nonnegative Matrix Factorization (HNMFk) and Boolean matrix factorization (BNMFk) with automatic model selection, as well as Logistic matrix factorization (LMF), we use to construct a three-level topic tree from a 46,862-document corpus focused on 73 transition-metal dichalcogenides (TMDs). This class of materials has been studied in a variety of physics fields and has a multitude of current and potential applications. 

An ensemble BNMFk + LMF approach fuses discrete interpretability with probabilistic scoring. The resulting HNMFk clusters map each material onto coherent research themes, such as superconductivity, energy storage, and tribology, and highlight missing or weakly connected links between topics and materials, suggesting novel hypotheses for cross-disciplinary exploration. We validate our method by removing publications about superconductivity in well-known superconductors, and demonstrate that the model correctly predicts thier association with the superconducting TMD clusters. This highlights the ability of the method to find hidden connections in a graph of material to latent topic associations built from scientific literature. This is especially useful when examining a diverse corpus of scientific documents covering the same class of phenomena or materials but originating from distinct communities and perspectives. The inferred links generating new hypotheses, produced by our method, are exposed through an interactive Streamlit dashboard, designed for human-in-the-loop scientific discovery.
\end{abstract}



 \maketitle

\section{Introduction}
Artificial intelligence is transforming materials science by addressing key challenges in discovery, synthesis, and property prediction. Recent advances, particularly in large language models (LLMs), transformers, and hybrid systems, have demonstrated strong performance in overcoming traditional bottlenecks \cite{liu2025large, van2025assessment}.

Here we present a novel, model-agnostic framework for missing link prediction between latent features extracted from a domain-specific ontology of 46,862 scientific papers on the 73 known transition metal dichalcogenides (TMDs) \cite{bhattarai2024healhierarchicalembeddingalignment}. The proposed workflow integrates:
a) the BUNIE method \cite{solovyev2023interactive} to build a targeted TMDs ontology,
b) HNMFk \cite{eren2023semi} with automatic topic number selection and interactive visualization tool, to extract latent topics, and
c) a custom Boolean Matrix Factorization with model selection, combined with Logistic Matrix Factorization (BNMFk-LMF) \cite{barron2025matrix}, to identify missing links in the \textbf{Materials Property Matrix}, derived form the TMDs corpus and HNMFk decomposition\footnote{NMFk, HNMFk, BNMFk, and LMF are available at \url{https://github.com/lanl/T-ELF} \cite{TELF}.}.

Our approach constructs a binary \textbf{Materials Property Matrix} from the extracted topics and material references, enabling prediction of functional links. We show that the proposed approach can effectively distinguish true superconductors from chemically similar non-superconductors and prioritize candidates for experimental validation. Its generalizable architecture supports broad application across material families and incomplete relational datasets, accelerating data-driven scientific discovery.
\section{Related Work}
Matrix completion algorithms have been specifically developed for discrete rating-scale data, addressing challenges such as corrupted observations and missing entries commonly encountered in recommender systems, surveys, and other partially observed datasets \cite{candes2010matrix}. Missing link prediction and matrix completion are closely related tasks, as both aim to recover missing values in incomplete matrices or graphs by exploiting underlying structure \cite{pech2017link, dunlavy2011temporal}. In the context of link prediction, the adjacency matrix of a graph is typically assumed to be low-rank, enabling the use of matrix completion techniques, such as low-rank approximation or factorization, to estimate the likelihood of unobserved connections \cite{kazemi2018simple}. Classical non-negative matrix factorization (NMF)–based link scoring continues to serve as a basis approach in this domain \cite{chen2017}.

For binary (categorical) networks with $\{0,1\}$ edge structures, missing link prediction seeks to infer unobserved relationships based on observed binary interactions. Logistic Matrix Factorization (LMF) models the probability of a link by applying a sigmoid transformation to the dot product of latent embeddings \cite{menon2011link, liu2016neighborhood}, whereas Boolean Matrix Factorization (BMF) directly approximates the binary adjacency matrix using logical operations such as AND and OR, preserving interpretability in discrete domains \cite{ravanbakhsh2016boolean}. For predicting material–property associations, \cite{sourati2023accelerating} use a hypergraph of articles combining entities and author collaborations, using embeddings and human-pathway-based transition probabilities. In contrast, our approach performs link prediction over latent topic–material associations to infer material–property relationships.
\section{Methods}
\subsection{Hierarchical Non‐negative Matrix Factorization (HNMFk)}
\label{sec:hnmfk}
Starting from the full matrix \(X\) and an index set \(I^{(0)}=\{1,\dots,n\}\),
HNMFk recursively applies NMFk to ever-smaller submatrices:

\begin{enumerate}[leftmargin=*,nosep]
  \item \textbf{Local factorisation.}  
        From the current index set \(I^{(d)}\) build  
        \(X^{(d)}\) (rows or columns), predict the stable rank \(k_d\) with
        NMFk, and factorise \(X^{(d)}\approx W^{(d)}H^{(d)}\).
  \item \textbf{Cluster assignment.}  
        Assign each sample \(i\in I^{(d)}\) to  
        \(\ell_i=\arg\max_r W^{(d)}_{ir}\) (or \(H^{(d)}\)), yielding \(k_d\) clusters.
  \item \textbf{Stopping test.}  
        Halt if any condition holds:  
        \(\min\!\dim X^{(d)}\le 1\); \(|I^{(d)}|\le s_{\min}\);
        \(k_d=1\); all \(\ell_i\) identical; or depth \(d=d_{\max}\).
  \item \textbf{Recursion.}  
        For every non-empty cluster \(c\) set  
        \(I^{(d+1)}_c=\{i\in I^{(d)}\mid\ell_i=c\}\), choose  
        \[
          K^{(d+1)}=\bigl\{k_{\min},\dots,
          \allowbreak\min(k_{\max},\,k_d+1)\bigr\},
        \]
        and repeat steps~1–4 on \(X^{(d+1)}_c\).
\end{enumerate}

By selecting \(k_d\) adaptively and stopping automatically, HNMFk
reveals a multiscale hierarchy with no manual tuning.

\subsection{Logistic Matrix Factorization}
\label{sec:lmf}

Logistic Matrix Factorization (LMF) extends classical low-rank factorization to {\em binary} interaction data.  
Given a labelled adjacency matrix \(X\in\{0,1\}^{n\times m}\), with \(X_{ij}=1\) if a link between entities
\(i\) and \(j\) is observed and \(X_{ij}=0\) otherwise, and a binary mask
\(M\in\{0,1\}^{n\times m}\) that distinguishes observed from unobserved entries,
LMF models the Bernoulli likelihood  
\begin{equation}
  \hat X \;=\; \sigma\!\bigl(W H + b_r + b_c\bigr), 
  \qquad \sigma(x)=\frac{1}{1+e^{-x}},
\end{equation}  
where \(W\in\mathbb{R}^{n\times k}\) and \(H\in\mathbb{R}^{k\times m}\) are latent
feature matrices, and \(b_r\in\mathbb{R}^{n\times 1}\), \(b_c\in\mathbb{R}^{1\times
m}\) are row- and column-specific bias vectors that capture node-level
propensities to form links.  
The parameters are learned by minimising the regularised negative
log-likelihood  
\begin{align}
  \mathcal{L}(W,H,b_r,b_c)
  &= -\!\!\sum_{i=1}^{n}\sum_{j=1}^{m}\!M_{ij}
     \Bigl[X_{ij}\log\hat X_{ij} + (1-X_{ij})\log(1-\hat X_{ij})\Bigr] \nonumber\\
  &\quad +\lambda\!\left(\lVert W\rVert_F^{2}+\lVert H\rVert_F^{2}
        +\lVert b_r\rVert_2^{2}+\lVert b_c\rVert_2^{2}\right),
\end{align}
which has simple gradient updates. The logistic link outputs calibrated probabilities so LMF is suitable link ranking \ref{sec:bnmfk_lmf}.

\subsection{\texorpdfstring{BNMFk-LMF}{BNMFk\_LMF}}
\label{sec:bnmfk_lmf}

BNMFk is the Boolean counterpart of NMFk, enforcing $\{0, 1\}$ constraints on the
factor matrices via adaptive thresholding while
automatically choosing the rank \(k\) through the NMFk stability heuristic.
We extended BNMFk with logistic ensembling (for details, see \cite{barron2025matrix}) to obtain {\bfseries BNMFk-LMF}. The
algorithm is as follows:

\begin{enumerate}
  \item Apply BNMFk to the topic-material matrix \(T\) to obtain the optimal rank \(k\) and the Boolean
        reconstruction \(\hat T = W H\).
  \item Fixing \(k\), run an LMF decomposition of \(T\) to learn row and column
        biases \(b_r, b_c\).
  \item Combine the two models and pass through the sigmoid:
        \begin{equation}
          \tilde T_{\text{final}}
          \;=\;
          \sigma\!\bigl(\hat T + b_r + b_c\bigr).
          \label{eq:bnmfk_lmf}
        \end{equation}
\end{enumerate}
Equation~\eqref{eq:bnmfk_lmf} unites BNMFk’s discrete structure discovery with LMF’s calibrated probabilities.

\section{Results}

\subsection{Materials Property Matrix}
\label{sec:data_materials}
The TMDs corpus, comprises 46,862 peer-reviewed articles.  
TMDs, with the general formula $MX_2$ ($M$ = transition metal, $X$ = chalcogen), are foundational to 2D materials research, known for their exceptional electronic, optical, and mechanical properties \cite{gupta2025two}. Despite only 72 known compounds, the combinatorial space of possible TMDs is vast and remains largely underexplored.

For hierarchical modeling we convert titles + abstracts of the selected by BUNIE scientific texts to TF–IDF matrix and
factorize it with HNMFk. A depth of \textbf{three} levels (\(L=3\))
proved optimal by bootstrap stability analysis, yielding coarse “super-topics”
at the root and progressively finer subtopics at lower depths.

We developed an interactive GUI for validating and exploring HNMF\textsubscript{k} document clusters. The interface consists of a \emph{query sidebar} and a \emph{results panel}, with an optional collapsible tree (not shown) that lets users select one or more HNMF\textsubscript{k} leaf clusters.

\begin{enumerate}[label=\textbf{\arabic*.},leftmargin=1.4em,nosep]
  \item \textbf{Sidebar (left).}  
        \begin{itemize}[nosep]
          \item \emph{Token Search.} Configure the number of visible ranked tokens per node (\texttt{Top n}; here \emph{All}), choose between \texttt{Keywords} and free-text \texttt{Denovo} modes, enter terms (e.g., \texttt{superconduct}), and set \texttt{and}/\texttt{or} logic.  
          \item \emph{Attribute Search.} Use dropdowns to apply metadata filters (country, author, affiliation, material, \ldots).  
          \item \emph{Cluster Picker.} A collapsible tree lists all HNMF\textsubscript{k} leaves; checking or unchecking boxes updates the working set dynamically.
        \end{itemize}
  \item \textbf{Results panel (right).}  
        Shows the current cluster (e.g., \texttt{18\_5\_0}) with a file-explorer link. Tabs provide summary views; the \texttt{Attributes} tab (shown) visualizes the \texttt{material} field for the 252 documents matching the query.  
        NbSe\textsubscript{2} dominates (67.9\%), followed by Se\textsubscript{2}Ti (8.9\%), CoS\textsubscript{2} and NbS\textsubscript{2} (5.4\% each), and smaller categories ($\le$ 3.6\%). Hovering shows counts; clicking a slice updates the cluster list, document table, and downstream charts.
\end{enumerate}

\noindent
The interface enables users to: (i) zoom into any part of the HNMF\textsubscript{k} hierarchy, (ii) apply rich metadata filters, and (iii) drill down from aggregate plots to individual documents, all within a unified view.

Using the TMDs corpus and the topics extracted via HNMFk, we construct the binary \textbf{Materials Property Matrix}. This matrix has 815 rows for to the discovered latent topics, and 72 columns representing the known $MX_2$ TMDs materials identified within the documents. Each entry $M_{ij}$ in the matrix is defined as follows: $M_{ij} = 1$ indicates that the $j^{\text{th}}$ material is associated with the $i^{\text{th}}$ topic, $M_{ij} = 0$ denotes no such association, and a missing value (NaN) implies insufficient information to determine the relationship.

\begin{figure}[ht]
    \vspace{-.9em}
    \centering
    \includegraphics[width=.8\columnwidth]{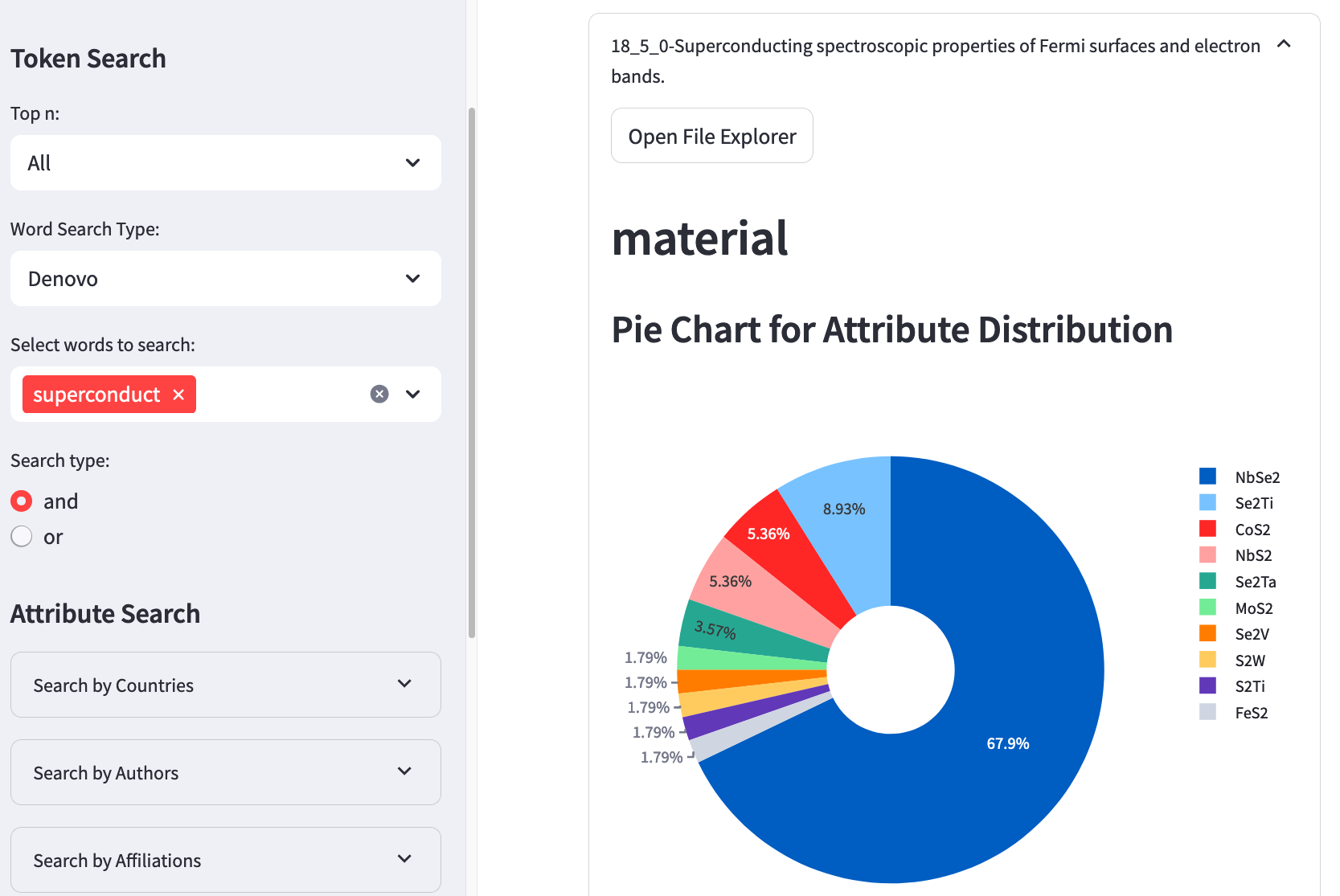}
    \vspace{-1.5em}
\caption{Dashboard used to explore the
HNMF\textsubscript{k} hierarchy.  }
    \vspace{-1em}
    \label{fig:frontend}
\end{figure}

\begin{figure}[ht]
    \vspace{-.9em}
    \centering
    \includegraphics[width=.8\columnwidth]{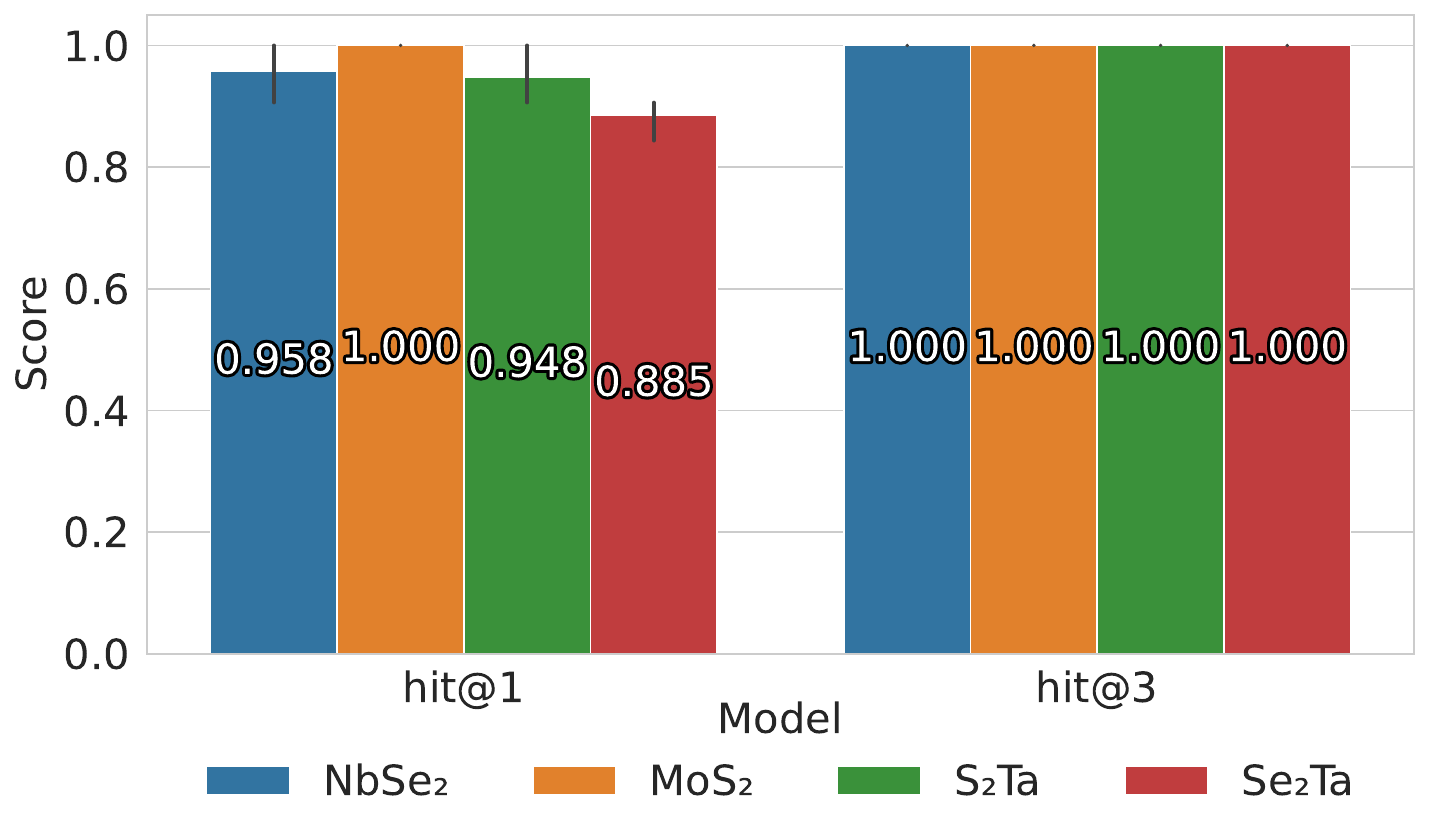}
    \vspace{-1.5em}
    \caption{Mean hit@1 and hit@3 scores (bars, annotated) of the
\textsc{BNMF}$+$\textsc{LMF} ensemble after masking every known
superconducting link and an equal number of random zeros for each of
the four benchmark compounds }
    \vspace{-1em}
    \label{fig:hit_scores}
\end{figure}

\subsection{Top \(k\) retrieval of hidden links}
\label{sec:topk_retrieval}

To evaluate whether our \(\textsc{BNMF}_k + \textsc{LMF}\) ensemble can \emph{rediscover} true but withheld superconducting links, we perform a leave-out experiment on four benchmark transition-metal dichalcogenides: NbSe\(_2\), MoS\(_2\), S\(_2\)Ta, and Se\(_2\)Ta.

\begin{enumerate}[label=(\arabic*)]
  \item \textbf{Masking.} For each compound, remove all verified superconducting links from the compound–superconductor cluster matrix, and randomly sample an equal number of non-superconducting (zero) entries as negatives.
  \item \textbf{Training and scoring.} Train the ensemble on the masked matrix and compute predicted scores \(s_{ij}\) for each masked entry \((i,j)\).
  \item \textbf{Metric.} Define
    \[
      \mathrm{hit@}k
      \;=\;
      \frac{1}{\lvert \mathcal P\rvert}
      \sum_{(i,j)\in\mathcal P}
      \mathbf{1}\bigl\{\mathrm{rank}(s_{ij}) \le k\bigr\},
    \]
    where \(\mathcal P\) is the set of masked positives, and \(\mathrm{rank}(s_{ij})\) is the score's position among all masked entries sorted in descending order.
  \item \textbf{Cross-validation.} Repeat the above over three random splits to estimate 95\% confidence intervals.
\end{enumerate}

Figure~\ref{fig:hit_scores} shows mean \(\mathrm{hit@1}\) and \(\mathrm{hit@3}\) across folds. Key findings:

\begin{itemize}
  \item \(\mathrm{hit@1} > 0.885\) for all compounds,
  \item \(\mathrm{hit@1} = 1.000\) for MoS\(_2\),
  \item \(\mathrm{hit@3} = 1.000\) for all four.
\end{itemize}

The narrow 95\% intervals confirm high consistency across folds. These results demonstrate that our ensemble reliably rediscovers masked superconducting links with near-perfect top-3 precision.

\subsection{Separation of positive and negative classes}

\begin{figure}[h]
  \centering
  \includegraphics[width=0.8\columnwidth]{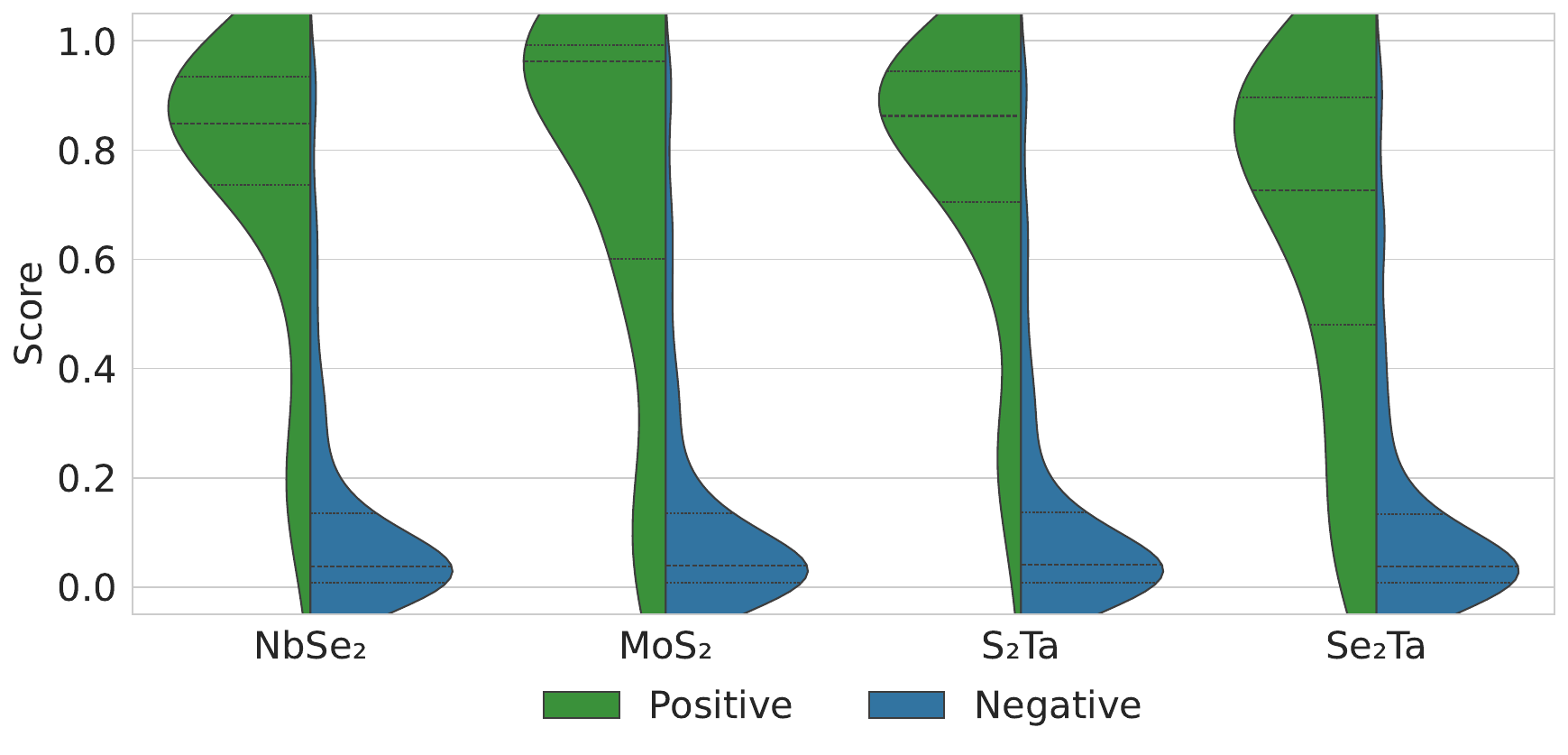}
  \caption{Split-violin plot of posterior probabilities for 24 superconducting edges (green). [See also Table \ref{tab:selected_scores}.] 192 non-superconducting edges (blue) were added for comparison. (Note that these are identical for the four violins.)  Lines denote the median and 25th/75th percentiles, with numbers showing cumulative counts. Superconductors concentrate near 1, non-superconductors near 0, indicating a clear separation.}
  \label{fig:half_violins}
\end{figure}

Figure~\ref{fig:half_violins} contrasts the posterior scores assigned to the
24 withheld positive edges (green violins) with the 192 masked
negatives (blue).  Positive edges cluster tightly near~1: the median sits above
0.90 and the lower quartile rarely drops below~0.85.  Negatives are strongly
skewed toward zero (median~$<\!0.05$); fewer of the 192 instances
per material penetrate the upper quartile.  The cumulative counts printed on
each quartile boundary emphasize the clean separation: most positives occupy
the top quartile, whereas the vast majority of negatives remain in the lowest.

\subsection{Ranking of unseen candidate materials}

Table~\ref{tab:selected_scores} reports the \emph{posterior} score assigned to each of the eleven compounds whose \textbf{labels were fully masked during training}.  In other words, every entry that would have indicated ``compound~$i$ is in a superconductor cluster’’ was removed from the observation matrix (\textit{masked as unknown}), so the model never saw a positive label for these materials while it was being fit.  The corresponding row and column themselves \emph{were} retained, populated only with zeros or missing values, so the model could still exploit their similarity to the rest of the data. 

\begin{table}
\caption{Average predicted scores for selected materials}
\label{tab:selected_scores}
    \scriptsize 
\begin{tabular}{l l r}
\toprule
Material & Set & Prediction Score \\
\midrule
S2Ta & Superconductor & 0.813 \\
NbSe2 & Superconductor & 0.757 \\
MoS2 & Superconductor & 0.703 \\
Se2Ta & Superconductor & 0.699 \\
FeS2 & Non-superconductor & 0.206 \\
MnS2 & Non-superconductor  & 0.116 \\
CrS2 & Non-superconductor  & 0.109 \\
MnSe2 & Non-superconductor  & 0.105 \\
MnTe2 & Non-superconductor  & 0.094 \\
CrSe2 & Non-superconductor  & 0.079 \\
CrTe2 & Non-superconductor  & 0.052 \\
\bottomrule
\end{tabular}
\end{table}

Importantly, the four layered dichalcogenides that are
\emph{experimentally} known to superconduct
(\emph{S\textsubscript{2}Ta}, \emph{NbSe\textsubscript{2}},
\emph{MoS\textsubscript{2}}, \emph{Se\textsubscript{2}Ta}) emerge at the very top of the ranking with scores in the range 0.70–0.81, whereas every material with no reported superconductivity receives a score $\le 0.206$.  The wide gap between the two groups implies that a single global threshold
($\approx 0.5$) would recover all known positives while rejecting all inspected negatives, demonstrating that the model can correctly infer superconducting behavior even when the ground-truth labels are hidden during training.

Taken together, the high hit@$k$ retrieval, the clear score
separation, and the compound‐level ranking show that the
the proposed approach: (i)\,rediscovers hidden superconducting links with essentially perfect precision and (ii)\,assigns suitably low confidence to chemically similar \emph{non}‐superconducting materials, highlighting its utility for guiding targeted experimental exploration.
\section{Conclusion}
Using BUNIE, HNMFk, and BNMFk–LMF, we analyzed a corpus spanning 73 known TMDs and built a sparse binary Material–Property Matrix (materials vs. hidden topics). We masked known superconducting and non-superconducting TMDs from the matrix and tested recovery without access to the withheld labels. Our model:

\begin{itemize}
    \item recovered all masked superconductors within the top three predictions (\(\mathrm{hit@3} = 1.00\)), with ~90\,\% ranked first,
    \item assigned high scores to positives (median~$\approx 0.91$) and low scores to most negatives (<~0.20), and
    \item ranked all four superconductors above all non-superconductors in a 15-compound test set, with a >0.5 score gap.
\end{itemize}

This model-agnostic workflow reliably distinguishes true superconductors from similar compounds and is transferable to other material families or incomplete relational datasets.

\begin{acks}
This research was funded by the U.S. Department of Energy National Nuclear Security Administration’s Office of Defense Nuclear Nonproliferation Research and Development (DNN R\&D), supported by the U.S. DOE NNSA under Contract No. 89233218CNA000001, as well as by the LANL Institutional Computing Program.
\end{acks}
\bibliographystyle{ACM-Reference-Format}
\bibliography{sections/ref}
\end{document}